\pdfoutput=1

\documentclass[letterpaper, 10pt, conference]{ieeeconf}  %

\IEEEoverridecommandlockouts                              %

\let\labelindent\relax                                    %

\usepackage{graphicx}            %
\usepackage{dblfloatfix}         %
\usepackage{amsmath}             %
\usepackage{amssymb}             %
\usepackage{amsfonts}            %
\usepackage{enumitem}            %
\usepackage{multirow}            %
\usepackage{siunitx}             %
\usepackage{url}                 %
\usepackage{xspace}              %
\usepackage[T1]{fontenc}         %
\usepackage[hidelinks]{hyperref} %

\makeatletter
\let\MYcaption\@makecaption
\makeatother
\usepackage[font=footnotesize]{subcaption}
\makeatletter
\let\@makecaption\MYcaption
\makeatother

\graphicspath{{images/}}
\DeclareGraphicsExtensions{.pdf,.png,.jpg,.jpeg}

\DeclareMathOperator{\sgn}{sgn}

\DeclareMathOperator{\coerce}{coerce}

\providecommand{\abs}[1]{\lvert#1\rvert}

\newcommand{\seclabel}[1]{\label{sec:#1}}

\newcommand{\figlabel}[1]{\label{fig:#1}}

\newcommand{\eqnlabel}[1]{\label{eqn:#1}}

\newcommand{\secref}[1]{Section~\ref{sec:#1}\xspace}

\newcommand{\figref}[1]{Fig.~\ref{fig:#1}\xspace}

\newcommand{\eqnref}[1]{(\ref{eqn:#1})\xspace}

\newcommand{\iguhop}{igus\textsuperscript{\tiny\circledR}$\!$ Humanoid Open Platform\xspace}
\newcommand{\iguhopp}{igus\textsuperscript{\tiny\circledR} Humanoid Open Platform\xspace}

\title{\LARGE \bf Omnidirectional Bipedal Walking with Direct\\Fused Angle Feedback Mechanisms}

\author{Philipp Allgeuer and Sven Behnke%
\thanks{All authors are with the Autonomous Intelligent Systems (AIS) Group, Computer Science Institute VI,
        University of Bonn, Germany. Email: {\tt\small pallgeuer@ais.uni-bonn.de}. This work was partially
        funded by grant BE 2556/10 of the German Research Foundation (DFG).}}

\usepackage{eso-pic}

\AtBeginDocument{\AddToShipoutPictureFG*{\AtTextUpperLeft{\put(0,\LenToUnit{9pt}){\parbox{\textwidth}{\centering\bfseries
International Conference on Humanoid Robots (Humanoids), Canc\'un, Mexico, 2016
}}}}}
\begin{document}

\bstctlcite{IEEEexample:BSTcontrol}

\maketitle
\thispagestyle{empty}
\pagestyle{empty}

\begin{abstract}
An omnidirectional closed-loop gait based on the direct feedback of orientation 
deviation estimates is presented in this paper. At the core of the gait is an 
open-loop central pattern generator. The orientation feedback is derived from a 
3D nonlinear attitude estimator, and split into the relevant angular deviations 
in the sagittal and lateral planes using the concept of fused angles. These 
angular deviations from expected are used by a number of independent feedback 
mechanisms, including one that controls timing, to perform stabilising 
corrective actions. The tuning of the feedback mechanisms is discussed, 
including an LQR-based approach for tuning the transient sagittal response. The 
actuator control scheme and robot pose representations in use are also 
addressed. Experimental results on an \iguhop demonstrate the core concept of 
this paper, that if the sensor management and feedback chains are carefully 
constructed, comparatively simple model-free and robot-agnostic feedback 
mechanisms can successfully stabilise a generic bipedal gait.
\end{abstract}

\section{Introduction}
\seclabel{introduction}

Walking is a fundamental skill for bipedal humanoid robots, but despite much 
research, is not yet considered solved. Walking poses many difficulties, 
including incomplete information, sensor delays, sensor noise, imperfect 
actuation, joint backlash, structural non-rigidity, uneven surfaces, and 
external disturbances.
The robot size is also a challenge, as larger robots often cannot move as 
quickly and easily as smaller ones, due to their generally lower torque to 
weight ratios. Larger robots are therefore more limited in the actions they can 
perform, and can often not use the same approaches to walking that smaller 
robots such as e.g.\ the Aldebaran Nao can, due to a violation of underlying 
dynamic and kinematic assumptions. Many approaches have been developed to target 
such problems, but frequently only for expensive high performance robots. 
Problems typical of low cost robots and sensors are not addressed in these 
works. Hence, the porting of such methods to cheaper bipedal robotic systems 
is often met with complication. This paper presents a method for flexible and 
reliable omnidirectional walking that is suitable for larger robots with low 
cost actuators and sensors, using the concept of direct fused angle feedback 
(see \figref{teaser}). It is demonstrated that genuinely good walking results 
can be achieved using relatively simple feedback mechanisms, with only minimal 
modelling of the robot, and without measuring or controlling forces or torques. 
Only joint positions and a 6-axis IMU are used. The contributions of this paper 
are the entire fused angle feedback scheme, the method of calculating 
feed-forward torques for position controlled gait applications, the listed 
extensions to the open-loop gait, and the described IMU calibration routines.

\begin{figure}[!t]
\parbox{\linewidth}{\centering\includegraphics[width=1.0\linewidth]{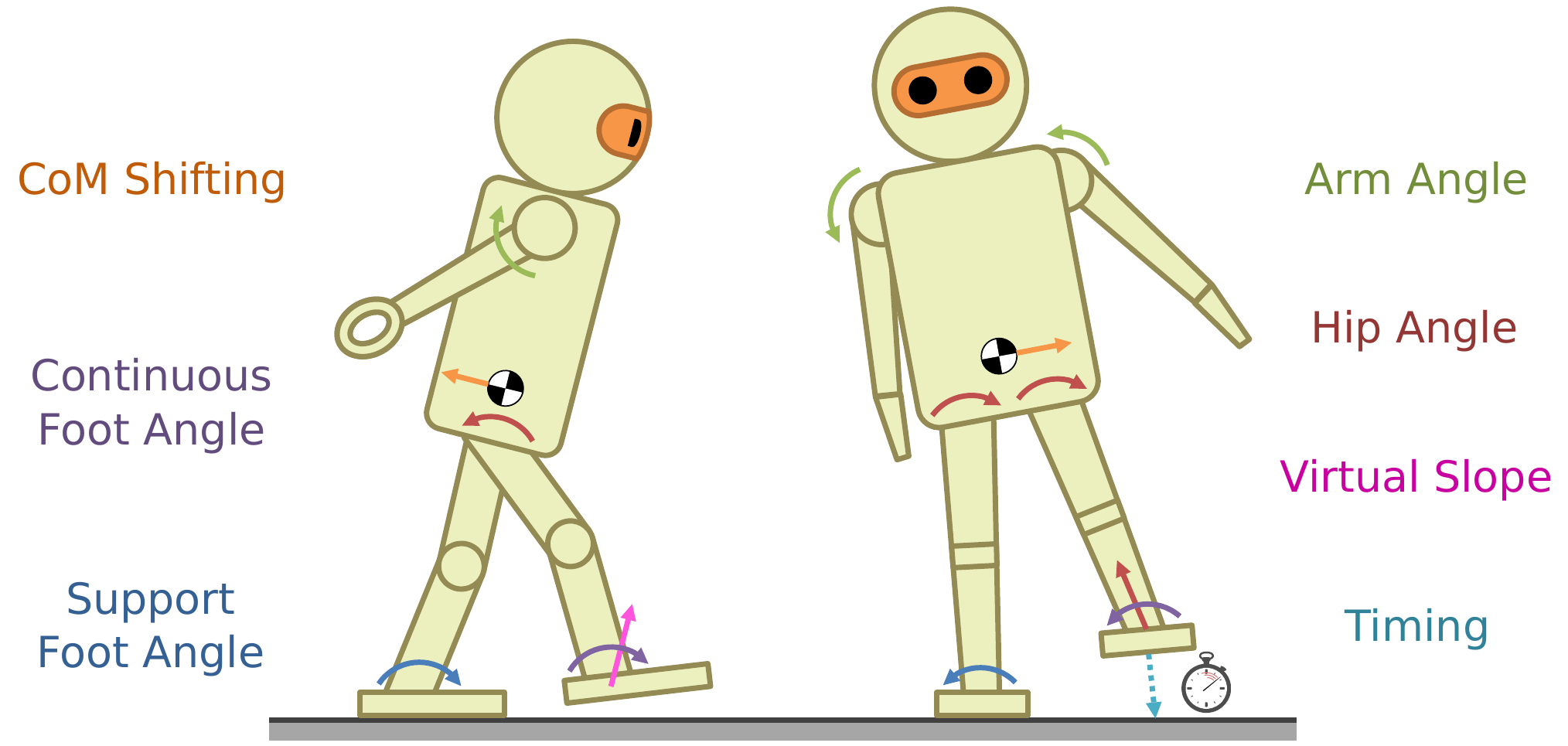}}
\caption{Example fused angle feedback corrective actions, including arm, hip, foot
and foot height components, in both the sagittal (left) and lateral (right) planes.
Step timing is also considered based on the lateral motion.}
\figlabel{teaser}
\vspace{-1ex}
\end{figure}
\section{Related Work}
\seclabel{related_work}

Many modern bipedal gaits are, in a wide variety of ways, based on the concept 
of the Zero Moment Point (ZMP). Given a piecewise polynomial ZMP trajectory, 
Harada et al. \cite{Harada2004} analytically derived a formulation for the 
reference centre of mass (CoM) trajectory. This was later extended by Morisawa 
et al. \cite{Morisawa2007} to allow modifications to foot placement. A basis for 
many works is ZMP tracking with preview control, first introduced by Kajita et 
al. \cite{Kajita2003}. A series of footsteps are planned and used to define a 
reference ZMP trajectory with the use of suitable heuristics. This trajectory is 
often locally fixed once it has been computed, but often also recomputed online 
in response to tracking errors and disturbances \cite{Tedrake2015}. A dynamic 
model of the robot is then used to optimise a smooth centre of mass (CoM) 
trajectory, e.g. in a model predictive control setting, that is maximally 
consistent with the planned ZMP trajectory \cite{Wieber2006}. If non-negligible 
CoM tracking errors occur, recomputation of the motion plan is required. This 
generally involves updating the footstep plan, which can be performed using a 
lower dimensional model, or by directly incorporating the footstep locations in 
the optimisation process \cite{Morisawa2010}. Kryczka et al. \cite{Kryczka2015} 
presented a ZMP-based method for incorporating, amongst other things, some level 
of timing feedback into the optimisation process. Their approach also requires 
measurement of the ZMP, and optimises a gait pattern depending on the Linear 
Inverted Pendulum Model (LIPM). Whenever the CoM feedback control is unable to 
regulate the CoM position to within a certain threshold of the reference 
trajectory, a recomputation of the gait pattern is triggered. Timing adjustment 
is only supported for lateral pushes towards the stance foot however.

When a disturbance to a stationary robot is detected, Urata et al. 
\cite{Urata2011} optimise two future footstep locations and one footstep time. 
The associated CoM trajectories are calculated explicitly using singular LQ 
preview regulation, based on the corresponding heuristically generated ZMP 
reference. The executed motion plan is fixed to always consist of exactly three 
steps, but this is sufficient to produce convincing disturbance rejection, 
albeit on specially designed hardware that is able to execute the planned 
motions with high fidelity.

It is a general problem of ZMP preview control gaits that good sensor feedback 
and CoM tracking performance are required, limiting the applicability of such 
approaches to high quality hardware, and/or smaller robots that have favourable 
torque to weight ratios. Furthermore, such gaits generally require impact force 
estimation, ZMP measurement through force-torque sensors, and/or precise 
physical modelling to guarantee the balance of the planned motions. These 
criteria are often not met with lower cost robots.

A distinct approach to balanced walking has been presented by Missura et al. in 
the form of the Capture Step Framework \cite{Missura2015}. This adjusts both the 
step position and timing to preserve balance, based on the prediction of the CoM 
trajectory using the LIPM. The main advantage of this method is that it does not 
require forces, torques or the ZMP to be measured, making it suitable for lower 
cost robots. One of the objectives of the gait proposed in this paper is to be 
able to serve as a stabilising foundation for more complex balancing schemes 
such as the capture steps.

\section{Open-Loop Walking}
\seclabel{open_loop_walking}

The open-loop gait core used in this work is based on the one proposed by 
Missura et al. \cite{Missura2013a}, but with numerous modifications and 
improvements. It is based on a central pattern generator that is able to make 
the \iguhop walk to some extent, but not as reliably as desired. Other robots 
with higher quality hardware, greater torque to weight ratios, less backlash 
and/or parallel leg kinematics can be made to walk more successfully with only 
an open-loop gait, but the feedback mechanisms are in each case still of great 
stability benefit.

\subsection{Actuator Control Scheme}
\seclabel{actuator_control}

The gait, as with most task-oriented motions of the \iguhopp, is relatively 
sensitive to how well the actuators track their set position. This is influenced 
by many factors, including battery voltage, joint friction, inertia, and the 
relative loadings of the legs. Feed-forward control is applied to the commanded 
servo positions to try to compensate all of these factors, with the aim of 
making the servo performance as consistent as possible across all the possible 
ranges of conditions. This allows the joints to be operated in higher ranges of 
compliance, reduces servo overheating and wear, increases battery life, and 
reduces the problems posed by impacts and disturbances.

Each actuator uses pure proportional position control. Given the current 
position $q$, and the desired setpoint $q_d$, it is assumed that the servo will 
produce a torque $\tau$ of
\begin{equation}
\tau = K_c V_B K_p (q_d - q), \eqnlabel{servo_torque}
\end{equation}
where $K_c$ is a constant proportional to the motor torque constant, and $V_B$ 
is the battery voltage \cite{Schwarz2013a}. Given a desired output torque 
$\tau_d$, and accounting for static, Coulomb and viscous friction, the total 
required torque is given by
\begin{equation}
\tau = \tau_d + \alpha_1 \dot{q} + \alpha_2 \sgn(\dot{q}) (1 - \beta) + \alpha_3 \sgn(\dot{q}) \beta,
\end{equation}
where $\alpha_{1,2,3}$ are constants of the motor, and $\beta \equiv 
\beta(\dot{q})$ characterises the Stribeck friction curve \cite{frictionmodel}. 
Thus, to produce the output torque $\tau_d$, the required position setpoint 
$q_d$ is
\begin{equation}
q_d = q + \tfrac{1}{V_B K_p}(\hat{\alpha}_0 \tau_d + \hat{\alpha}_1 \dot{q} + \hat{\alpha}_2 s_{\dot{q}} (1 - \beta) + \hat{\alpha}_3 s_{\dot{q}} \beta),
\end{equation}
where $s_{\dot{q}} \equiv \sgn(\dot{q})$, and $\hat{\alpha}_{0,1,2,3}$ are 
constants of the motor that incorporate $K_c$ from \eqnref{servo_torque}. The 
$\hat{\alpha}_{0,1,2,3}$ parameters were learnt from experiments using iterative 
learning control to maximise tracking performance \cite{Schwarz2013a}. The 
vector of desired feed-forward output torques $\boldsymbol{\tau}\!_d$ is computed 
from the vectors of commanded joint positions, velocities and accelerations 
using the full-body inverse dynamics of the robot, with help of the Rigid Body 
Dynamics Library. This is the only point where a model of the robot is required, 
and is optional. Setting all feed-forward torques to zero would only sacrifice 
the gravity and robot inertia compensation, but still leave the friction, 
battery voltage and compliance compensation intact.

A so-called \emph{single support model} is created for the trunk, and for each 
link that is at the tip of a limb. It is assumed in each of these dynamics 
models that the corresponding root link for which it was created, is fixed in 
free space, and that no other links have external contacts. This allows inverse 
dynamics computations to be made. In addition to the commanded positions, 
velocities and accelerations, it is assumed in every time step that a set of 
support coefficients are commanded, one for each single support model, e.g. from 
the gait. The coefficients are normalised to sum to one across all root links, 
and express the proportion of the robot's weight that is expected to be carried 
by each root link. The inverse dynamics of the trunk single support model is 
computed first, with only the commanded positions, velocities and accelerations 
applied, to estimate the inertial torques required to produce the commanded 
accelerations. The inverse dynamics is then calculated for each single support 
model in turn, modelling only the application of a gravity force to the model, 
and the resulting sets of joint torques are averaged together by the principle 
of superposition, using the commanded support coefficients as weights. As a 
necessary simplification for stability reasons, the gravity force is always 
assumed to point in a fixed direction relative to the trunk in every support 
model. Perhaps unintuitively, the error in this assumption is actually less than 
if orientation feedback were to be used, because of the complex nature of ground 
contacts. The resulting torques are added to those calculated from the inertial 
components to give the final desired feed-forward output torques 
$\boldsymbol{\tau}\!_d$.

\subsection{The Joint, Abstract and Inverse Pose Spaces}
\seclabel{pose_spaces}

Three spaces are used to represent the pose of the robot---the joint space, 
abstract space and inverse space. The \emph{joint space} pose is defined as the 
vector $\mathbf{q}$ of all joint angles. The \emph{inverse space} pose is given 
by the vector containing the Cartesian coordinates and quaternion orientations 
of each of the limb end effectors relative to the trunk link frame. The sign 
convention used in this work is that the x-axis points forwards, the y-axis 
points left, and the z-axis points upwards. The last of the three 
representations, the \emph{abstract space}, is a representation that was 
specifically developed for humanoid robots in the context of walking and 
balancing \cite{Behnke2006}. The abstract space reduces the expression of the 
pose of each limb to parameters that define the length of the limb, the 
orientation of a so-called limb centre line, and the orientation of the end 
effector. The abstract leg parameters can be computed from the joint angles 
$\mathbf{q}$ as,
\begin{align}
\eta &= 1 - \cos(\tfrac{1}{2} q^{knee}_{pitch}), & \xi_{Lz} &= q^{hip}_{yaw}, \\
\xi_{Ly} &= q^{hip}_{pitch} + \tfrac{1}{2} q^{knee}_{pitch}, & \xi_{Lx} &= q^{hip}_{roll}, \\
\xi_{Fy} &= \xi_{Ly} + q^{ankle}_{pitch} + \tfrac{1}{2} q^{knee}_{pitch}, & \xi_{Fx} &= \xi_{Lx} + q^{ankle}_{roll},
\end{align}
where $\eta$ is the \emph{leg extension}, $\xi_{Lx}, \xi_{Ly}, \xi_{Lz}$ are the 
\emph{leg angles}, and $\xi_{Fx}, \xi_{Fy}$ are the \emph{foot angles}. 
Analogous formulas exist to define the \emph{arm extension} and \emph{arm 
angles}. The complete abstract pose representation is then the combination of 
the abstract pose parameters for each arm and leg. Simple conversions between 
all three pose spaces exist, with the only required kinematic knowledge being 
the limb link lengths, for conversions to and from the inverse space.

\begin{figure*}[!t]
\parbox{\linewidth}{\centering
\includegraphics[height=36.7mm]{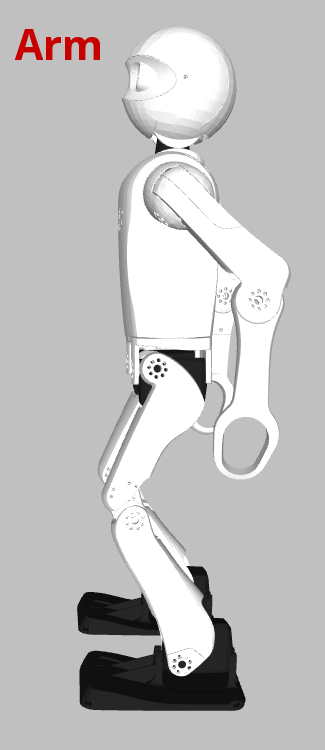}%
\includegraphics[height=36.7mm]{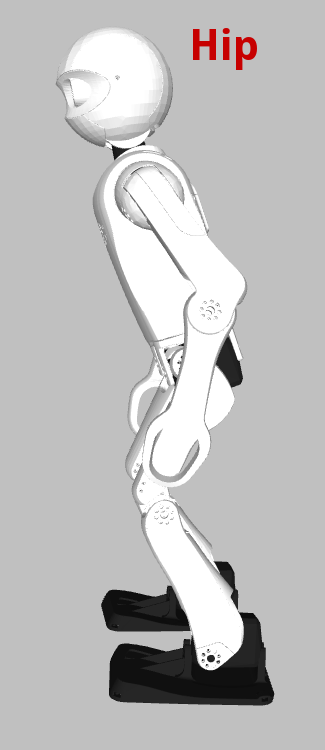}%
\includegraphics[height=36.7mm]{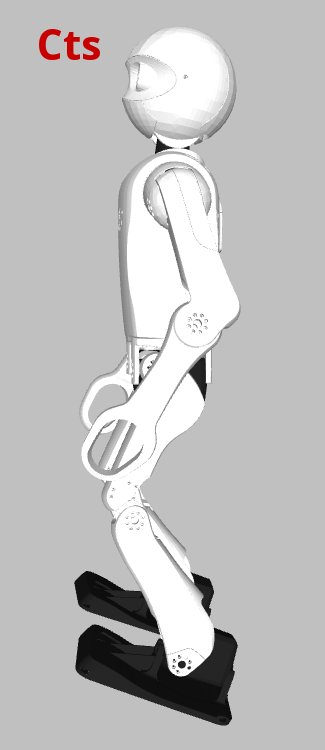}%
\includegraphics[height=36.7mm]{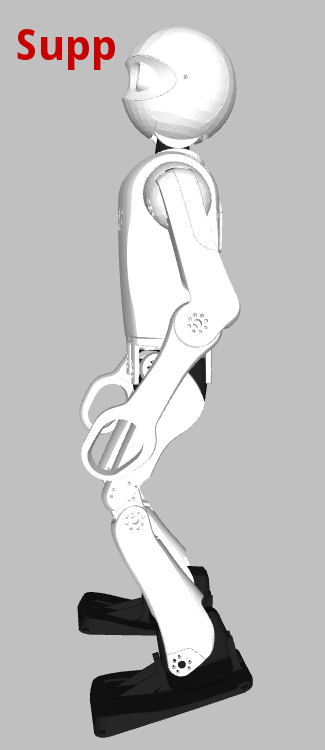}%
\includegraphics[height=36.7mm]{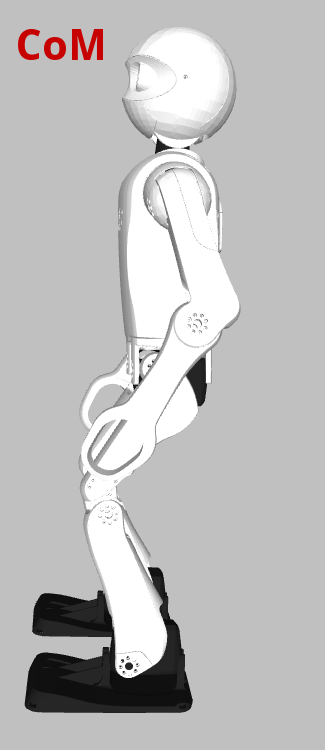}\hspace{3pt}%
\includegraphics[height=36.7mm]{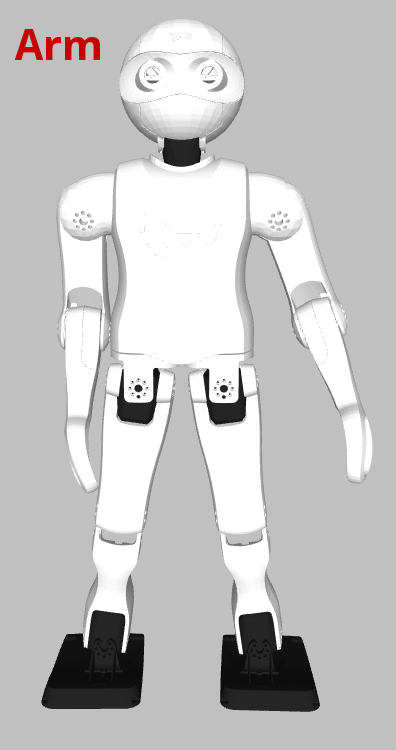}%
\includegraphics[height=36.7mm]{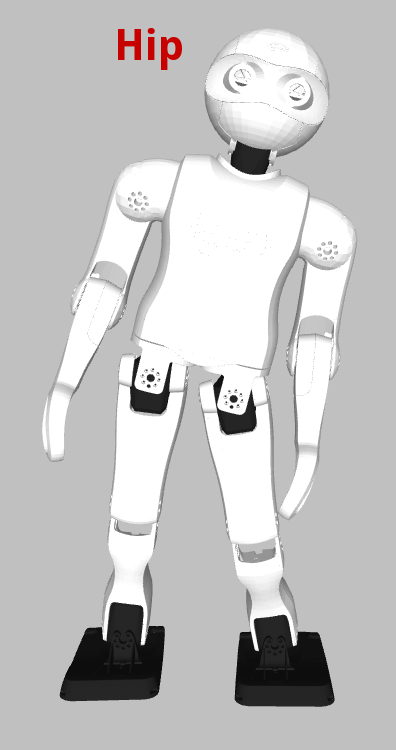}%
\includegraphics[height=36.7mm]{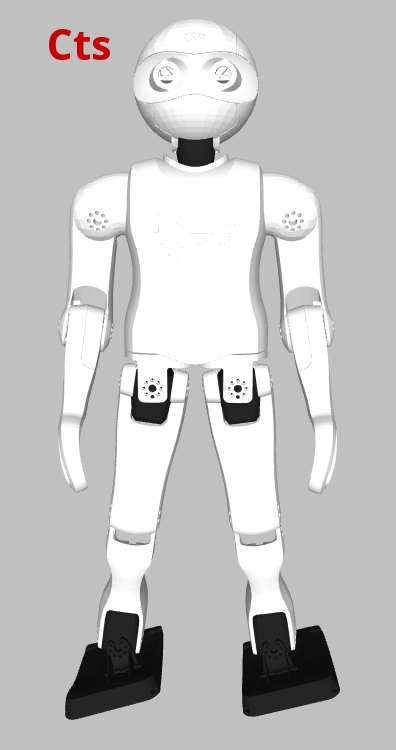}%
\includegraphics[height=36.7mm]{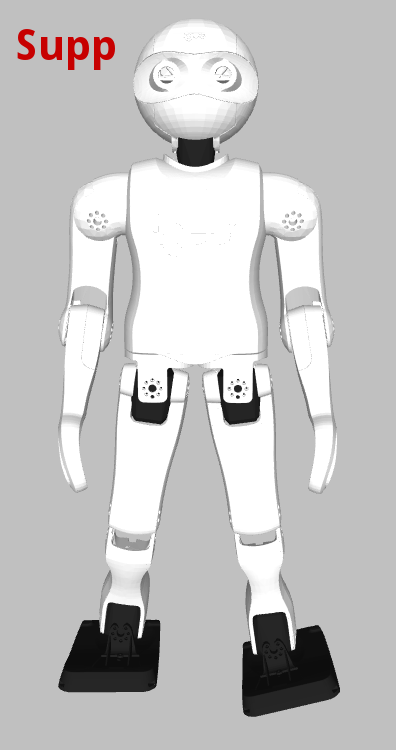}%
\includegraphics[height=36.7mm]{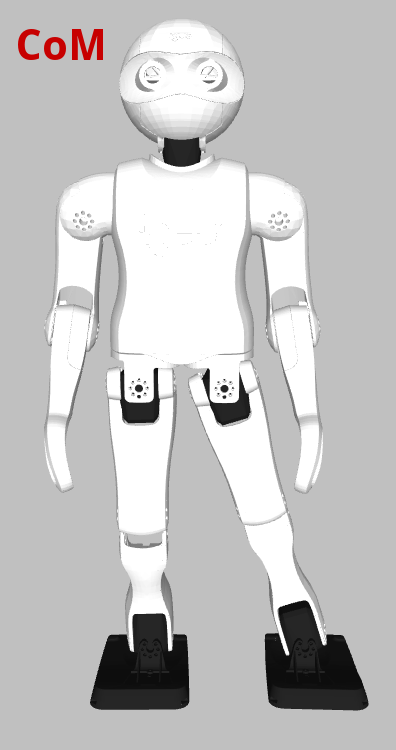}}
\caption{The implemented corrective actions in both the sagittal (left image) 
and lateral (right image) planes, from left to right in both cases the arm 
angle, hip angle, continuous foot angle, support foot angle, and CoM shifting 
corrective actions. The actions have been exaggerated for clearer illustration.}
\figlabel{corrective_actions}
\vspace{-2ex}
\end{figure*}
\subsection{Central Pattern Generated Gait}
\seclabel{cpg_gait}

The open-loop gait used in this paper significantly extends the one from previous work 
\cite{Missura2013a}. The main changes include:
\begin{itemize}
\item Changes to the leg extension profiles to transition more smoothly between 
swing and support phases,
\item The addition of a double support phase for greater walking stability and 
passive oscillation damping,
\item The addition of a trim factor for the angle relative to the ground at 
which the feet are lifted during stepping,
\item The integration of a dynamic pose blending algorithm to enable smoother 
transitions to and from walking,
\item The incorporation of support coefficient waveforms, for use with the 
actuator control scheme,
\item The introduction of a leaning strategy based on the rate of change of the 
commanded gait velocity, and
\item The use of hip motions instead of leg angle motions for the gait command 
velocity-based leaning strategies.
\end{itemize}

The central pattern generator at the core of the open-loop gait produces the 
high-dimensional trajectories required for omnidirectional walking in a mix of 
the three pose spaces. The pose at each point in time is a deterministic 
function of the commanded gait velocity $\mathbf{v}_g$, which is \mbox{norm-,} 
acceleration- and jerk-limited for stability reasons, and the \emph{gait phase} 
$\mu \in (-\pi,\pi]$. The gait phase starts at 0 or $\pi$, and is updated after 
each time step $\Delta{}t$ using
\begin{equation}
\mu[n+1] = \mu[n] + \pi f_g \Delta{}t, \eqnlabel{gait_phase_update}
\end{equation}
where $f_g$ is the configured gait frequency in units of steps per second. By 
convention, $\mu = 0$ and $\mu = \pi$ correspond to the commanded touchdown 
times of the individual legs.

The central pattern generated gait begins with a configured halt pose in the 
abstract space, then incorporates numerous additive waveforms to the halt pose 
as functions of $\mathbf{v}_g$ and $\mu$. These waveforms generate features such 
as leg lifting, leg swinging, arm swinging, and so on. The resulting abstract 
pose is then converted to the inverse space, where further motion components are 
added. The resulting inverse pose is finally converted to the joint space, in 
which form it is commanded to the robot via the actuator control scheme 
described in \secref{actuator_control}. The fused angle feedback mechanisms work 
by adding extra components to the central pattern generated waveforms, in both 
the abstract and inverse spaces.

\section{State Estimation}
\seclabel{state_estimation}

The main source of feedback for the closed-loop gait is the orientation of the 
robot, as it encodes to a great extent the state of balance of the robot. This 
is estimated from 3-axis gyroscope and accelerometer data, using the attitude 
estimator from \cite{Allgeuer2014}. The resulting orientation is split into its 
components in the sagittal and lateral planes using \emph{fused angles}, to 
allow independent feedback loops to be applied in the sagittal and lateral 
directions. Fused angles, developed by Allgeuer et al. \cite{Allgeuer2015}, 
are a novel way of representing orientations that offer significant benefits 
over Euler angles for applications that involve balance. The resulting estimated 
fused pitch $\theta_B$ and fused roll $\phi_B$ of the robot form the basis of 
the feedback for the closed-loop gait.

Although the attitude estimator continuously estimates the gyroscope bias, the 
high level of noise in the measured accelerations limits the gain with which 
this can be done. Three calibration routines have been implemented for the 
gyroscope and accelerometer to maximise the quality of the fused angle 
estimates. High and low temperature gyro scale calibrations ensure through 
saturated linear interpolation that the angular velocities provided by the 
gyroscope are accurate in magnitude. The calibration is done by running the 
estimation while rotating the robot in an unbiased way through large known 
angles, and providing the appropriate calibration triggers. An orientation 
offset calibration compensates for any differences in orientation between the 
trunk link frame and the frame of the inertial sensors. Finally, a gyroscope 
bias auto-calibration is run online to complement the bias estimation that is 
performed in the attitude estimator. Brief intervals where the robot is at rest 
are automatically detected and used to converge the gyroscope bias estimate to 
the true measured value. This automatically provides a good initial estimate of 
the gyro bias, and is robust to sudden changes in sensor conditions.

\section{Fused Angle Feedback Mechanisms}
\seclabel{basic_feedback}

Feedback mechanisms have been implemented that apply direct additive corrective 
actions to the gait, based on the deviations of the fused pitch $\theta_B$ and 
fused roll $\phi_B$ from their nominal limit cycle values. The corrective 
actions can be thought of as motion primitives that are dynamically weighted and 
superimposed on top of the open-loop gait. The limit cycles of the fused angles 
are modelled as parametrised sine functions of the gait phase, and the fused 
angle deviations $d_{\theta}, d_{\phi}$ are defined as the difference between 
$\theta_B, \phi_B$ and their current expected values. Corrective actions in both 
the abstract space and inverse space (see \secref{pose_spaces}) have been 
implemented, namely (see \figref{corrective_actions}):

\begin{itemize}
\item \textbf{Arm angle:} The abstract arm angles are adjusted to bias the 
robot's balance, and produce reaction moments that help counterbalance transient 
instabilities (e.g. moving the arms backwards tilts the robot backwards).

\item \textbf{Hip angle:} The torso of the robot is tilted within the lateral 
and sagittal planes to induce lean in a particular direction, with adjustment of 
the abstract leg extension parameters preventing a counterproductive ensuing 
difference in foot height (e.g. leaning towards the left makes the robot swing 
out more to the left).

\item \textbf{Continuous foot angle:} Continuous offsets are applied to the 
abstract foot angles to bias the tilt of the entire robot from the feet up (e.g. 
putting the front of the feet more down makes the robot lean more backwards).

\item \textbf{Support foot angle:} Gait phase-dependent offsets are applied to 
the abstract foot angles (e.g. tilting the inside support foot edge down induces 
greater centre of mass swing onto the support foot). The offsets are faded in 
linearly just after the corresponding leg is extended fully, and faded out 
linearly just before the leg begins to retract again for its swing phase. As 
such, the offsets are applied only during the respective expected support 
phases, and at no point simultaneously.

\item \textbf{CoM shifting:} The inverse kinematic positions of the feet 
relative to the torso are adjusted in the horizontal plane to shift the position 
of the centre of mass (CoM), thereby adjusting the centering of the robot's mass 
above its support polygon (e.g. shifting the CoM to the right trims the time 
spent on each foot to the right).

\item \textbf{Virtual slope:} The inverse kinematic positions of the feet 
relative to the torso are adjusted in the vertical direction in a gait 
phase-dependent manner to lift the feet more at one swing extremity. This can be 
thought of as what the robot would need to do to walk up or down a slope. See 
\secref{virtual_slope} for details.
\end{itemize}

The complete feedback pipeline, from the fused angle deviations through to the 
resulting timing and corrective actions, is shown in \figref{pipeline}. The 
various feedback mechanisms, including the calculation 
of the fused feedback vector
\begin{equation}
\mathbf{e} = 
\begin{bmatrix}
e_{Px} & \!e_{Py} & \!e_{Ix} & \!e_{Iy} & \!e_{Dx} & \!e_{Dy}
\end{bmatrix}^T\!\!,
\eqnlabel{feedback_vector}
\end{equation}
are discussed in the following sections. Once $\mathbf{e}$ has 
been calculated, it is converted to a vector of corrective action activation 
values $\mathbf{u}_a$ via premultiplication by a corrective action gains matrix 
$K_a$. That is,
\begin{equation}
\mathbf{u}_a = K_a \mathbf{e}.
\end{equation}
Note that a single $5\times6$ gains matrix is used for generality and 
mathematical brevity, but in practice a full tune involves only 10-14 non-zero 
gains in the matrix. These intended feedback paths are explicitly stated in the following 
sections. The entries of $\mathbf{u}_a$ give the magnitude and sign with which 
each of the available corrective actions---excluding the virtual slope, which is 
calculated and added separately (see \secref{virtual_slope})---are applied to 
the abstract and inverse poses generated by the gait. This is how the robot acts 
to keep its balance.

To ensure that the pose of the robot stays within suitable joint limits, smooth 
saturation is applied to the final abstract arm angles, leg angles, foot angles, 
and inverse kinematic adjustments to the CoM. Smooth saturation, also referred 
to as \emph{soft coercion}, of a value $x$ to the range $(m,M)$ with a buffer of 
$0 < b \le \tfrac{1}{2}(M-m)$, is given by
\begin{equation}
\tilde{x} =
\begin{cases}
M - b e^{-\tfrac{1}{b}(x - (M - b))} & \text{if $x > M - b$}, \\
m + be^{\tfrac{1}{b}(x - (m + b))} & \text{if $x < m + b$}, \\
x & \text{otherwise}.
\end{cases}
\end{equation}
The soft coercion transfer function is shown in \figref{transfer_functions}.
Soft coercion is used instead of the \emph{standard coercion} $\coerce(x,m,M)$, 
which simply saturates $x$ to the interval $[m,M]$ in a binary manner, because 
soft coercion is continuously differentiable (class $\mathcal{C}^1$), resulting 
in smoother saturation behaviour and less self-disturbances of the robot.

\subsection{Fused Angle Deviation Proportional Feedback}
\seclabel{fused_feedback}

The most fundamental and direct form of fused angle feedback is the proportional 
corrective action feedback. As with nearly all of the implemented feedback 
mechanisms, the proportional feedback operates in both the lateral and sagittal 
planes. The fused angle deviation values $d_{\theta}, d_{\phi}$ are first passed 
through a mean filter to mitigate the effects of noise. Smooth deadband is then 
applied to the output to inhibit corrective actions when the robot is close to 
its intended trajectory, to avoid oscillations due to hunting. \emph{Smooth 
deadband} (see \figref{transfer_functions}) is a $\mathcal{C}^1$ adaptation of 
the standard `sharp' deadband, and for input $x$ and deadband radius $r$ is 
given by
\begin{equation}
\tilde{x} = 
\begin{cases}
\tfrac{\; x^2}{4r} \sgn(x) & \text{if $\abs{x} < 2r$}, \\
x - r \sgn(x) & \text{otherwise}.
\end{cases}
\end{equation}
Smooth deadband was chosen for this application to soften the transitions 
between action and inaction, and to avoid any unnecessary self-disturbances 
and/or unexpected oscillatory activation-deactivation cycles, which can occur 
due to the strongly asymmetrical behaviour of sharp deadband around its 
threshold points. The proportional fused angle feedback values $e_{Px}$ and 
$e_{Py}$ from \eqnref{feedback_vector} are calculated by scaling the results of 
the smooth deadband by the proportional fused angle deviation gain $K_p$. The 
proportional fused angle feedback mechanism is intended for activating the 
\emph{arm angle}, \emph{support foot angle}, and \emph{hip angle} corrective 
actions.

\begin{figure}[!t]
\parbox{\linewidth}{\centering\includegraphics[width=0.8\linewidth]{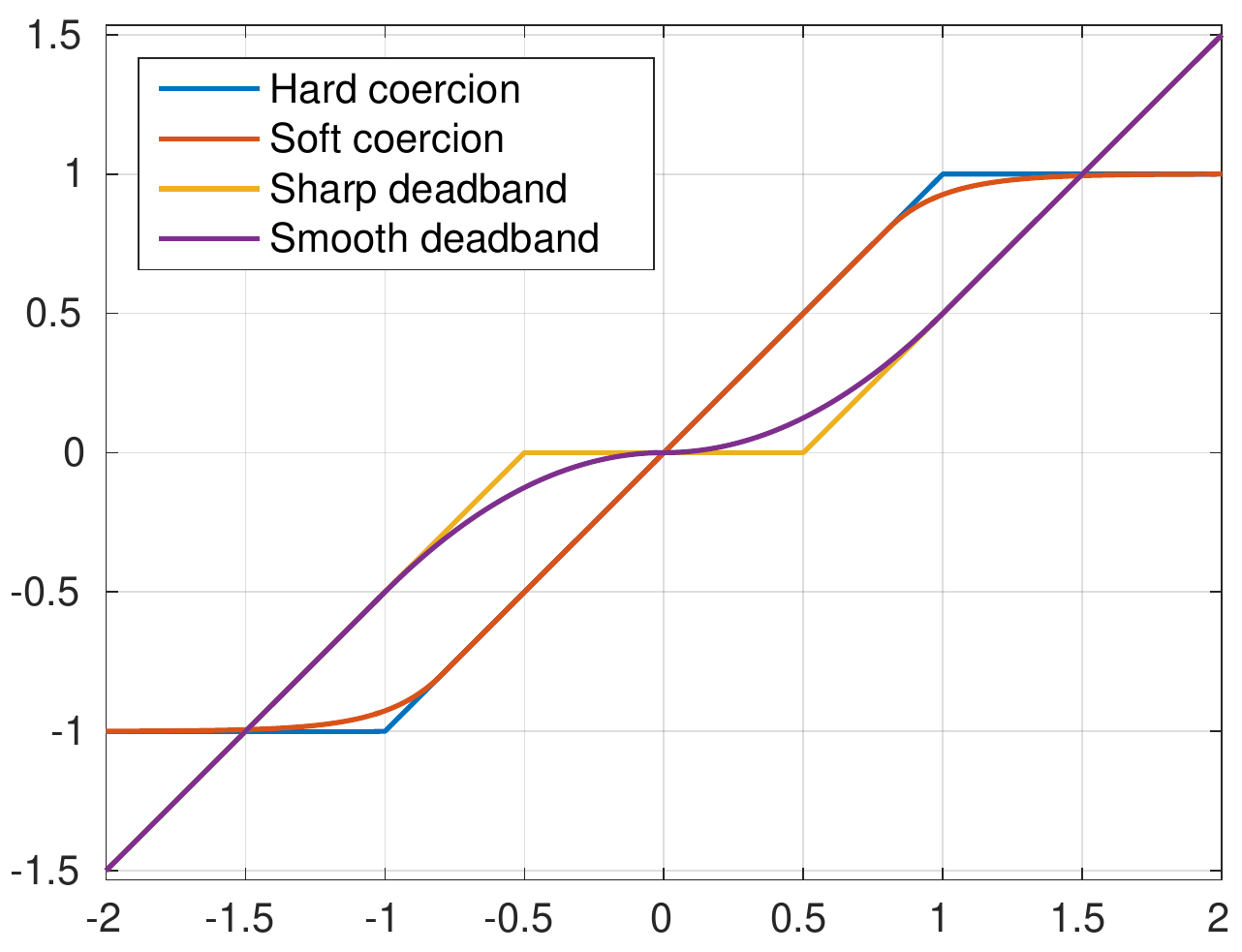}\hspace*{2mm}}
\caption{Transfer functions of the soft coercion and smooth deadband filters, shown in comparison to the hard coercion and sharp deadband filters.}
\figlabel{transfer_functions}
\vspace{-2ex}
\end{figure}
\subsection{Fused Angle Deviation Derivative Feedback}
\seclabel{dfused_feedback}

The proportional feedback works well in preventing falls when the robot is 
disturbed, but if used alone has the tendency to produce possibly unstable low 
frequency oscillations in the robot. To enhance the transient disturbance 
rejection performance of the robot, corresponding derivative feedback terms are 
incorporated to add damping to the system. If the robot has a non-zero angular 
velocity, this component of the feedback reacts `earlier' to cancel out the 
velocity before a large fused angle deviation ensues, and the proportional 
feedback has to combat the disturbance instead.

The derivative fused angle feedback values $e_{Dx}$ and $e_{Dy}$ are computed by 
passing the fused angle deviations $d_{\phi}$ and $d_{\theta}$ through a 
weighted line of best fit (WLBF) derivative filter, applying smooth deadband, 
and then scaling the results by the derivative fused angle deviation gain $K_d$. 
The smooth deadband, like for the proportional case, is to ensure that no 
corrective actions are taken if the robot is within a certain threshold of its 
normal walking limit cycle, and ensures that the transition from inaction to 
action and back is smooth.

\begin{figure}[!t]
\parbox{\linewidth}{\centering\includegraphics[width=1.0\linewidth]{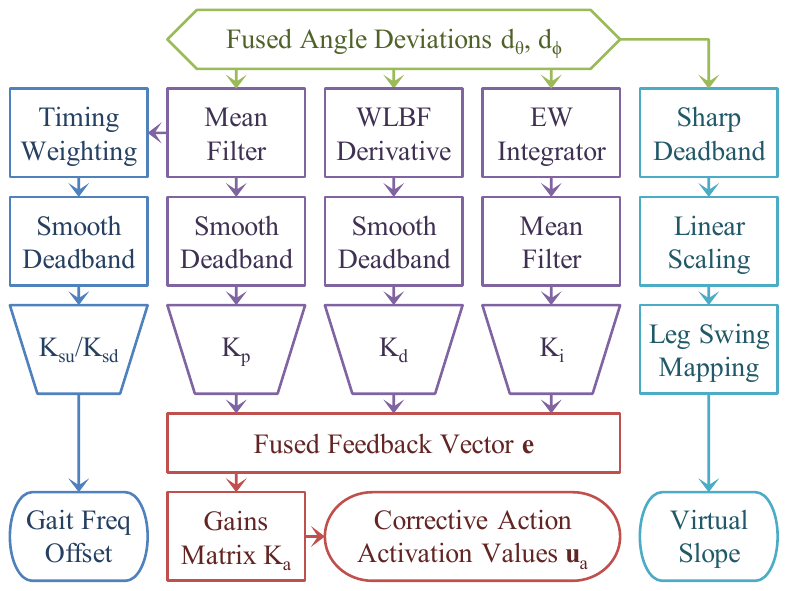}}
\caption{Overview of the complete fused angle feedback pipeline.}
\figlabel{pipeline}
\vspace{-2ex}
\end{figure}

A \emph{weighted line of best fit (WLBF) filter} observes the last $N$ data 
points of a signal, in addition to assigning confidence weights to each of the 
data points, and performs weighted linear least squares regression to fit a line 
to the data, with the data measurement timestamps being used as the independent 
variable. The linear fit evaluated at the current time gives a smoothed data 
estimate, and the fitted linear slope gives a smoothed estimate of the 
derivative of the data stream.
WLBF derivative filters have a number of advantages, in particular when compared 
to the alternative of computing the numerical derivative of a signal, and 
applying a low pass filter. WLBF filters are more robust to high frequency noise 
and outliers in the data for the same level of responsiveness to input 
transients, and have the advantage of inherently and easily supporting 
non-constant time separations between data points in a stable manner, even if 
two data points are very close in time. The use of weights in WLBF filters also 
allows some data points to be given higher confidences than others in a 
probabilistic manner. For example, in regular parts of the gait cycle where 
noise and errors in the measurements are expected, lower weights can be used to 
help reject noise and disturbances in the sensor measurements. To the knowledge 
of the authors, the use of weighted linear least squares regression in this 
online fashion for the purposes of data smoothing and derivative estimation, in 
particular in the context of walking gaits, has not previously been published, 
so no references can be provided.

The derivative fused angle feedback values $e_{Dx}$ and $e_{Dy}$, are intended 
for activating the \emph{arm angle}, \emph{support foot angle} and 
\emph{hip angle} corrective actions, and serve to allow better loop shaping for 
the transient response to disturbances.

\subsection{Fused Angle Deviation Integral Feedback}
\seclabel{ifused_feedback}

The proportional and derivative fused angle feedback mechanisms are able to 
produce significant improvements in walking stability, but situations may arise 
where continued corrective actions are required to stabilise the robot, due to 
for example asymmetries in the robot. The continual control effort, and 
resulting periodic steady state errors, are undesired. The purpose of the 
integral feedback is to slowly converge on offsets to the gait halt pose that 
minimise the need for control effort during general walking.

The fused angle deviations $d_{\phi}$ and $d_{\theta}$ are first integrated over 
time using an \emph{exponentially weighted (EW) integrator}, a type of `leaky 
integrator'. This computes the sum
\begin{equation}
I[n] = x[n] + \alpha x[n-1] + \alpha^2 x[n-2] + \ldots
\end{equation}
where $\alpha \in [0,1]$ is the history time constant. The sum is 
most conveniently computed using the difference equation
\begin{equation}
I[n] = x[n] + \alpha I[n-1].
\end{equation}
If $\alpha = 0$, the integrator simply returns the last data value, but if 
$\alpha = 1$ the output is the same as that of a classical integrator, so the value of $\alpha$ effectively trims the amount of memory that the 
integrator has. A suitable value for $\alpha$ is computed from the desired 
half-life time $T_h$, which is a measure of the decay time of the integrator, 
using
\begin{equation}
\alpha = 0.5^{\tfrac{\Delta{}t}{T_h}},
\end{equation}
where $\Delta{}t$ is the nominal time step. An EW integrator is used instead of 
a standard integrator for flexibility and stability reasons, to combat 
integrator windup, and because old data eventually needs to be `forgotten' while 
walking, because situations can change. The alternative of keeping the last $N$ 
fused angle deviations in a buffer, and integrating over the buffer, is less 
efficient, less continuous in the output and less flexible, in addition to being 
vulnerable to aliasing effects.

The outputs of the two fused angle deviation EW integrators are passed through 
mean filters, to allow the trade-off between settling time and level of noise 
rejection to be trimmed, and scaled by the integral fused angle deviation gain 
$K_i$ to give the integral fused angle feedback values $e_{Ix}$ and $e_{Iy}$. 
The integral fused angle deviation feedback is intended for activating a mix of 
the \emph{arm angle}, \emph{continuous foot angle}, \emph{hip angle}, and 
\emph{CoM shifting} corrective actions, although to preserve greater 
independence between the feedback mechanisms, primarily the last of these four 
is used.

\subsection{Fused Angle Deviation Timing Feedback}
\seclabel{timing_feedback}

The PID feedback mechanisms act to return the robot to the
expected fused angle limit cycles during walking. For larger disturbances this 
is not always possible, practicable and/or desired. For instance, if a robot is 
pushed so far laterally that it is only resting on the outer edge of its support 
foot, then no matter what fused angle feedback mechanisms are applied, it is 
infeasible to return the robot to its centred and upright position in the same 
time as during normal walking. This generally results in the robot attempting to 
make its next step anyway, and falling. To deal with this mode of failure, timing 
feedback has been implemented. This form of feedback adjusts the rate of 
progression of the gait phase as a function of the fused roll deviation 
$d_{\phi}$. Steps taken by the robot can thereby be sped up or delayed based on 
the lateral balance state of the robot.

As the required sign of the timing feedback depends on the current support leg, 
and as only minimal timing adjustments are desired during the double support 
phase due to the fused angle noise associated with foot strike and weight 
shifting, the mean filtered fused roll deviation $\tilde{d}_{\phi}$ is first 
weighted by a saturated oscillatory gait phase-dependent expression:
\begin{equation}
\hat{d}_{\phi} = \tilde{d}_{\phi} \coerce\bigl(-K_{tw}\sin(\mu - \tfrac{1}{2} \mu_{ds}),-1,1\bigr)
\end{equation}
where $\mu$ is the gait phase, $\mu_{ds}$ is the double support phase length, 
and $K_{tw} \in [1,\infty)$ is a gain that adjusts the shape of the oscillatory 
weighting curve. To avoid unnecessary disturbances to the gait frequency during 
normal walking, smooth deadband is applied to the calculated weighted deviation 
$\hat{d}_{\phi}$ to give the fused angle timing feedback value $e_{Tx}$. The 
desired gait frequency $f_g$ is then
\begin{align}
\tilde{f}_g &=
\begin{cases}
f_n + K_{su} e_{Tx} & \text{if $e_{Tx} \ge 0$}, \\
f_n + K_{sd} e_{Tx} & \text{otherwise},
\end{cases} \\
f_g &= \coerce(\tilde{f_g}, 0, f_{max}),
\end{align}
where $f_n$ is the nominal gait frequency, $f_{max}$ is the maximum allowed gait 
frequency, and $K_{su}$ and $K_{sd}$ are the speed-up and slow-down gains. 
Equation \eqnref{gait_phase_update} is then used as normal to update the gait 
phase $\mu$.

\subsection{Virtual Slope Walking}
\seclabel{virtual_slope}

It can happen that a foot prematurely strikes the ground during its swing phase 
if the robot is tilted sagittally in the direction that it is walking. This is 
undesirable, and is a cause for destabilisation of the robot. Virtual slope 
walking adjusts the inverse kinematic height of the feet in a gait 
phase-dependent manner, based on the commanded gait velocity and the estimated 
attitude, to simulate the robot walking up or down a slope. For example, if the 
robot is walking forwards and tilted forwards, the robot lifts its feet higher 
at the front of its swing phases, ensuring that the legs still reach full 
forward swing before establishing ground contact.

The required sagittal virtual slope $\theta_v$ at each point in time is 
calculated as the fused pitch deviation $d_{\theta}$ with sharp deadband 
applied, and asymmetrically linearly scaled based on whether the robot is tilted 
in the direction it is walking or not. The virtual slope $\theta_v$ is then 
scaled by the sagittal gait command velocity $v_{gx}$, and applied to the 
inverse kinematic foot height as a linear function of the corresponding current 
sagittal leg swing angle.

\section{Tuning of the Feedback Parameters}
\seclabel{tuning}

The process of tuning the feedback parameters is greatly simplified by the 
considerable independence of the feedback mechanisms. The individual feedback 
mechanisms also have clearly observable and direct actions that can be precisely 
isolated and tested, so arguably the process of tuning the proposed gait is 
quicker and easier than it would be for most model-based approaches that do not 
work out of the box.

The PD gains are tuned first, to establish the set of most effective corrective 
actions, and the gain ranges that produce noticeable effect without risking 
oscillations or instabilities. The choice of corrective actions may 
be influenced by external objectives, such as for example the desire to 
minimise trunk angle deviations from upright. In this case, the hip angle 
actions may for example be omitted. The transient sagittal response to 
disturbances comes predominantly from the PD feedback mechanisms, and can be 
tuned using an LQR-based approach. The robot is made to walk on the spot while 
activating the corrective actions with a multi-frequency signal, in a ratio 
chosen based on the relative desired ranges of the corrective actions. In the 
case of the \iguhop, the System Identification Toolbox in MATLAB was used to fit 
a generic second order state space system to the observed data, giving a model of
\begin{align}
\dot{x} &= A x + B u, & y &= C x + D u,\\
A &=
\begin{bmatrix}
-31.82 & 14.83 \\
207.5 & -221.3
\end{bmatrix}\!\!, &
B &=
\begin{bmatrix}
-51.67\\
720.2
\end{bmatrix}\!\!,\\
C &=
\begin{bmatrix}
1.185 & 0.1683
\end{bmatrix}\!\!, &
D &=
\begin{bmatrix}
0
\end{bmatrix}\!\!,
\end{align}
where the input $u$ is the activation of the corrective actions and the output 
$y = d_{\theta}$ is the fused pitch deviation. Cross-validation of the model was 
performed on two further data sets. The model has one fast pole 
(\SI{236.3}{\radian\per\second}), attributed to the immediate conservation of 
momentum effect of moving the arms, and one slower pole 
(\SI{16.77}{\radian\per\second}), which is approximately equal to the nominal 
gait frequency, and thus attributed to the gait and stepping dynamics. Similar 
to the approach in \cite{Mukhopadhyay1978}, the PD feedback law is given by
\begin{align}
u &= -K_p y - K_d \dot{y} = -K x, \\
K &= (\mathbb{I} + K_d C B)^{-1}(K_p C + K_d C A). \eqnlabel{K_defn}
\end{align}
A standard LQR design approach is then used to compute the gains vector $K$ that
minimises the performance cost function
\begin{equation}
J = \int_{0}^{\infty}\!\!x^T\!Q x + u^T\!R u\,dt.
\end{equation}
The $Q$ and $R$ matrices were initially chosen based on Bryson's rule 
\cite{Franklin2015} and subsequently refined. The maximum desired values of $x$ 
and $u$ for use with the rule were determined based on the simulated response of 
the model to the three data sets. Once $K$ has been calculated, the optimal PD 
gains, by inversion of \eqnref{K_defn}, are then
\begin{equation}
\begin{bmatrix}
K_p & \!\!K_d
\end{bmatrix}
= K
\begin{bmatrix}
C \\
C A - C B K
\end{bmatrix}^{-1}
\end{equation}
Due to modelling limitations, the final gains that arise will usually require 
some manual fine-tuning to get the most out of the feedback, often an increase 
in the D gain. Locally around upright the sagittal dynamics are somewhat stable, 
so the LQR naturally assumes this also holds at greater trunk attitudes, but 
this is not correct.

Once the transient response has been tuned, a suitable integral feedback 
half-life is chosen based on the rate at which the robot should adapt to a new 
environment, and gains are chosen that brings the activation of the associated 
integral corrective actions to the desired range. Timing is then considered, 
with the speed-up and slow-down gains being selected to avoid premature 
stepping, and to instantaneously halt the gait phase when the robot reaches a 
certain nominal lateral angular deviation. The virtual slope mechanism is mostly 
parameter-free, and the sole linear scaling factor is chosen to provide the 
desired margin of clearance of the foot from the ground during maximum forwards 
walking.

\section{Experimental Results}
\seclabel{experimental_results}

\begin{figure}[!t]
\parbox{\linewidth}{\centering
\includegraphics[width=\linewidth]{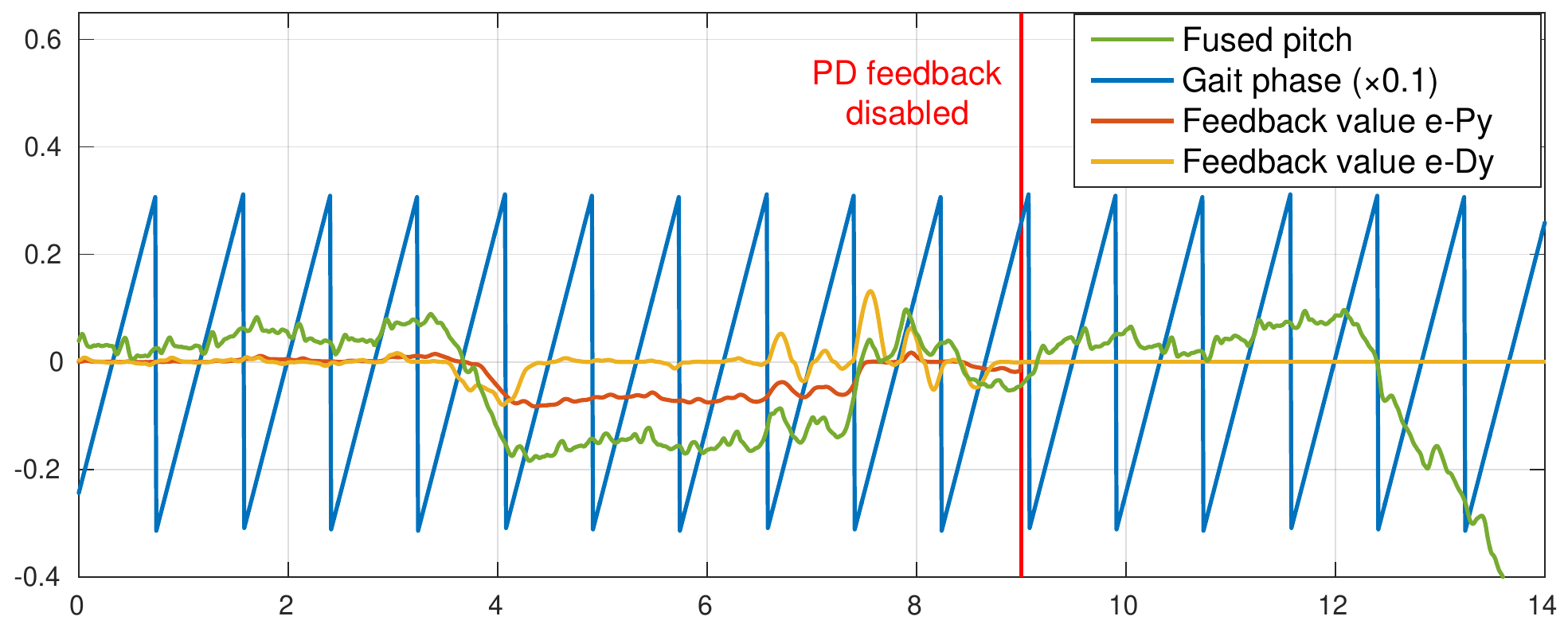}
\vspace{-4ex}\subcaption{The effect of the proportional and derivative fused angle feedback mechanisms when walking onto a step change in floor height.}\figlabel{result_walkonbook}
\vspace{1ex}\includegraphics[width=\linewidth]{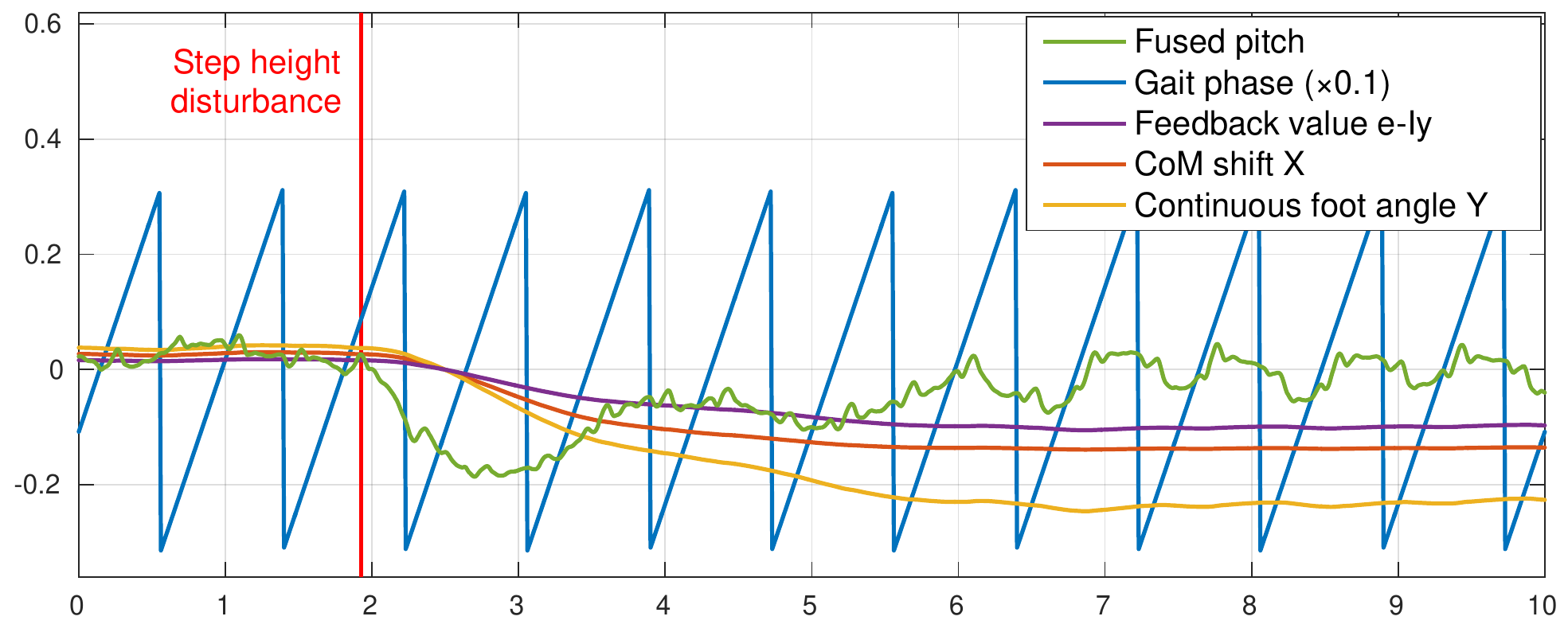}
\vspace{-4ex}\subcaption{The effect of the integral fused angle feedback mechanism when walking onto a step change in floor height.}\figlabel{result_learnbook}
\vspace{1ex}\includegraphics[width=\linewidth]{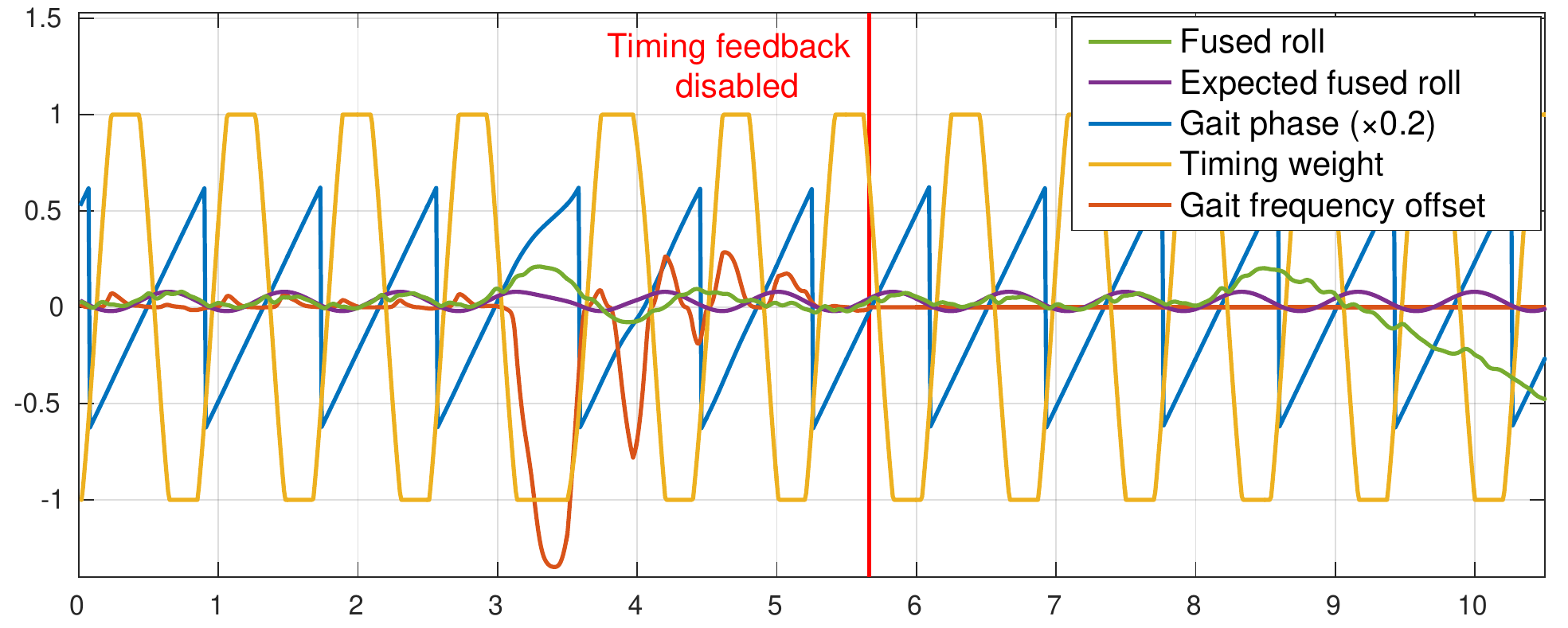}
\vspace{-4ex}\subcaption{The effect of timing feedback on lateral disturbance rejection.}\figlabel{result_timing}
\vspace{1ex}\includegraphics[width=\linewidth]{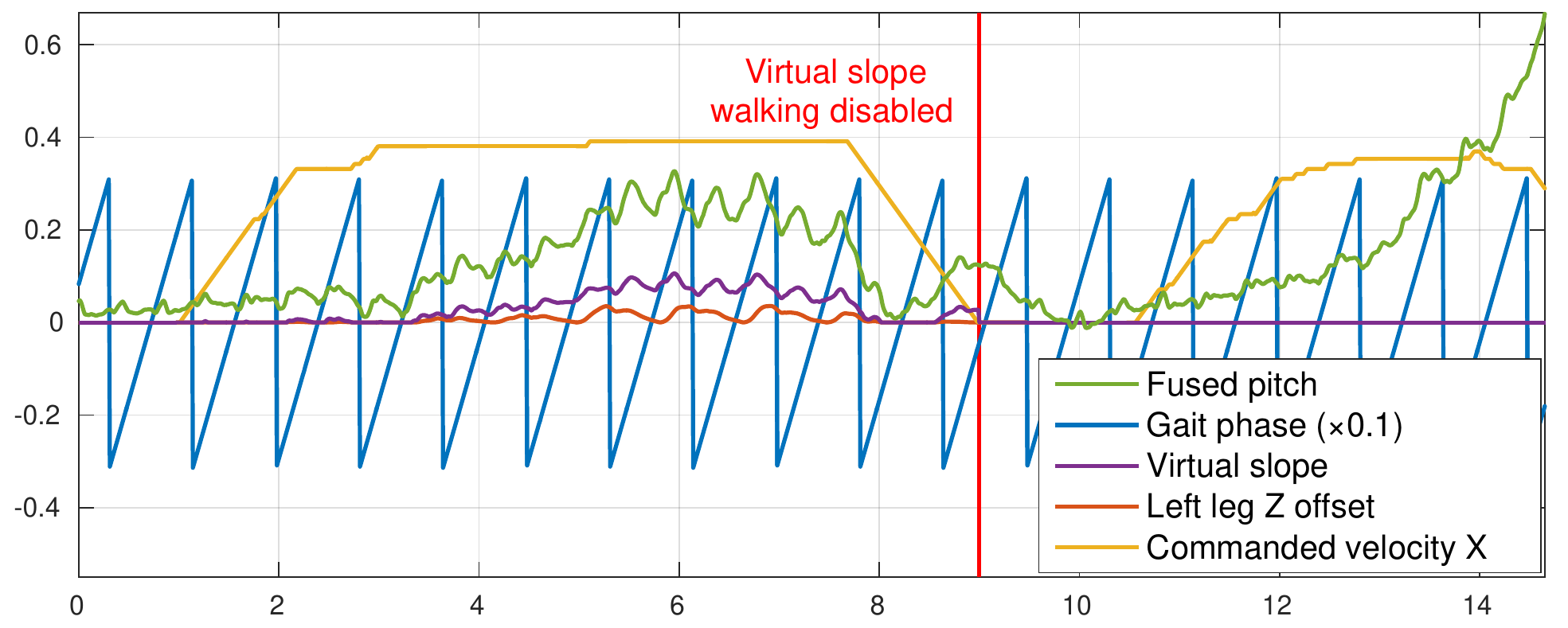}
\vspace{-4ex}\subcaption{The effect of the virtual slope on forwards walking while being pushed.}\figlabel{result_virtualslope}}
\caption{Results of the gait showing the effects of the feedback mechanisms.}
\figlabel{results}
\vspace{-1.5ex}
\end{figure}

The proposed gait has been implemented and evaluated experimentally on four 
\iguhop robots \cite{Allgeuer2015b}, and for cross-validation also on the 
humanoid robot Dynaped. The latter is dynamically quite different from the igus 
robots, but was able to use essentially the same gains matrix $K_a$. This 
demonstrates a strength of the gait, in that it is model-free and thereby 
relatively easy to transfer between robots. Overall, the feedback mechanisms 
were observed to make a significant difference in the walking ability of the 
tested robots, with walking often not even being possible for extended periods 
of time without them. The feedback mechanisms also imparted the robots with 
disturbance rejection capabilities that were not present otherwise. Reliable 
omnidirectional walking speeds of \SI{21}{\centi\metre\per\second} were achieved 
on an artificial grass surface of blade length \SI{32}{\milli\metre}.

Plots of experimental results demonstrating the efficacy of the feedback 
mechanisms are shown in \figref{results}. In \figref{result_walkonbook}, the 
robot walked twice onto an unexpected \SI{1.5}{\centi\metre} step change in 
floor height, the first time with proportional and derivative feedback enabled, 
and the second time with them disabled. It can be seen that with the feedback 
enabled the robot is able to avoid falling---albeit with a large steady state 
error in the fused pitch---while without feedback the robot falls immediately. 
Adding in integral feedback, configured to activate both the CoM shift and 
continuous foot angle corrective actions, produces the results in 
\figref{result_learnbook}. It can be seen that the relatively large error in the 
fused pitch is rejected in about \SI{5}{\second}, with the robot converging to 
upright walking despite the uneven floor. Note that the residual fused pitch 
limit cycles are an effect of the robot's feet not being equally positioned on 
the step.

\figref{result_timing} shows the effect of the timing feedback. The robot was 
subjected to two lateral pushes while walking in place, the first with timing 
feedback enabled, and the second without. It can be seen at $t = 
\SI{3.2}{\second}$ that the gait phase slows down in response to the push, 
allowing the fused roll to return to its expected limit cycle within 
\SI{2.5}{\second}. An identical push at $t = \SI{8.2}{\second}$, without timing 
feedback enabled, causes the robot to fall. \figref{result_virtualslope} 
illustrates the effect of virtual slope walking. The robot was continuously 
pushed forwards while it was walking forwards, first with the virtual slope 
walking enabled, and then without. In the first case, the adjustments to the 
inverse kinematic height of the feet ensured that the robot could continue to 
walk forwards, and regain balance once it was no longer being pushed. In the 
latter case however, the feet started to collide with the ground in front of the 
robot. The robot was not able to get its feet underneath its CoM again, and 
subsequently fell shortly after.

The gait presented in this paper has successfully been run in combination with 
the capture step framework proposed in \cite{Missura2015}, and produced results 
notably superior to just using the latter on the \iguhop. The capture step 
timing was used in place of the fused angle deviation timing, and the step sizes 
computed from the capture step algorithm were used, but otherwise the full fused 
angle feedback mechanisms were running. The gait was used in this configuration 
at the RoboCup 2016 soccer tournament, and over all games played, none of the 
five robots ever fell while walking\footnote{Video: 
\url{https://www.youtube.com/watch?v=G9llFqAwI-8}}, excluding cases of strong 
collisions with other robots. This validates the use of the presented gait as a 
stabilising foundation for more complex balancing schemes.

\section{Conclusion}
\seclabel{conclusion}

An inherently robust omnidirectional closed-loop gait has been presented that 
stabilises a central pattern generated open-loop gait using fused angle feedback 
mechanisms. The gait is simple, model-free, quick to tune, easily transferable 
between robots, and only requires servo position feedback if feed-forward torque 
compensation is desired. The gait is also suitable for larger robots with 
low-cost sensors and position-controlled actuators. This demonstrates that 
walking does not always mandate complex stabilisation mechanisms. The gait has 
been experimentally verified and discussed, and demonstrably made a robot walk 
that was not able to produce nearly similar results with just a manually tuned 
open-loop approach. One of the notable merits of the presented gait is that it 
can combine very well with more complicated model-based approaches that are able 
to suggest step size and/or timing adjustments. This is what makes the gait so 
useful and powerful as a building block for more complex and more tailored gait 
stabilisation schemes.

\bibliographystyle{IEEEtran}
\bibliography{IEEEabrv,ms}

\end{document}